\documentclass[10pt,twocolumn,letterpaper]{article}

\usepackage{cvpr}              
\usepackage{algorithm,algorithmic}
\usepackage{multirow}
\newcommand\blfootnote[1]{%
	\begingroup 
	\renewcommand\thefootnote{}\footnote{#1}%
	\addtocounter{footnote}{-1}%
	\endgroup 
}

\definecolor{cvprblue}{rgb}{0.21,0.49,0.74}
\usepackage[pagebackref,breaklinks,colorlinks,allcolors=cvprblue]{hyperref}

\title{Multi-party Collaborative Attention Control for Image Customization}

\author{Han Yang$^{1,2}$ \qquad Chuanguang Yang$^{1*}$ \qquad Qiuli Wang$^3$ \qquad Zhulin An$^{1*}$  \qquad Weilun Feng$^{1,2}$ \\
Libo Huang$^1$ \qquad Yongjun Xu$^1$ \\
$^1$Institute of Computing Technology, Chinese Academy of Sciences, Beijing, China  \\ 
$^2$University of Chinese Academy of Sciences, Beijing, China \\
$^3$Department of Radiology, The First Affiliated Hospital of Army Medical University, Chongqing, China \\
{\tt\small \{yanghan22s, yangchuanguang, anzhulin, huanglibo, fengweilun24s, xyj\}@ict.ac.cn} \\
{\tt\small \{wangqiuli@tmmu.edu.cn\}} \\
}

\begin{document}
\maketitle

\begin{abstract}
The rapid advancement of diffusion models has increased the need for customized image generation. However, current customization methods face several limitations: 1) typically accept either image or text conditions alone; 2) customization in complex visual scenarios often leads to subject leakage or confusion; 3) image-conditioned outputs tend to suffer from inconsistent backgrounds; and 4) high computational costs.
To address these issues, this paper introduces \textbf{M}ulti-party \textbf{C}ollaborative \textbf{A}ttention \textbf{C}on\textbf{tr}o\textbf{l} (MCA-Ctrl), a tuning-free method that enables high-quality image customization using both text and complex visual conditions. Specifically, MCA-Ctrl leverages two key operations within the self-attention layer to coordinate multiple parallel diffusion processes and guide the target image generation. This approach allows MCA-Ctrl to capture the content and appearance of specific subjects while maintaining semantic consistency with the conditional input.
Additionally, to mitigate subject leakage and confusion issues common in complex visual scenarios, we introduce a Subject Localization Module that extracts precise subject and editable image layers based on user instructions. Extensive quantitative and human evaluation experiments  show that MCA-Ctrl outperforms existing methods in zero-shot image customization, effectively resolving the mentioned issues.
\end{abstract}    
\section{Introduction}
\label{sec:intro}
\blfootnote{
$^*$ Corresponding author.

Code: https://github.com/yanghan-yh/MCA-Ctrl
}

Recent advances in generative artificial intelligence (GenAI) have greatly enhanced text-to-image (T2I) models \cite{rombach2022high, zhang2023adding, nichol2021glide, podell2023sdxl, ramesh2022hierarchical, hu2024instruct, feng2023ranni, feng2024relational, yang2024clip, feng2025mpq}, enabling them to generate realistic images from user prompts. As T2I models evolve, there has been an increasing demand for customized image creation \cite{gal2022image, ruiz2023dreambooth, kumari2023multi, li2023dreamedit}.

\begin{figure}[t]
  \centerline{\includegraphics[width=1\linewidth]{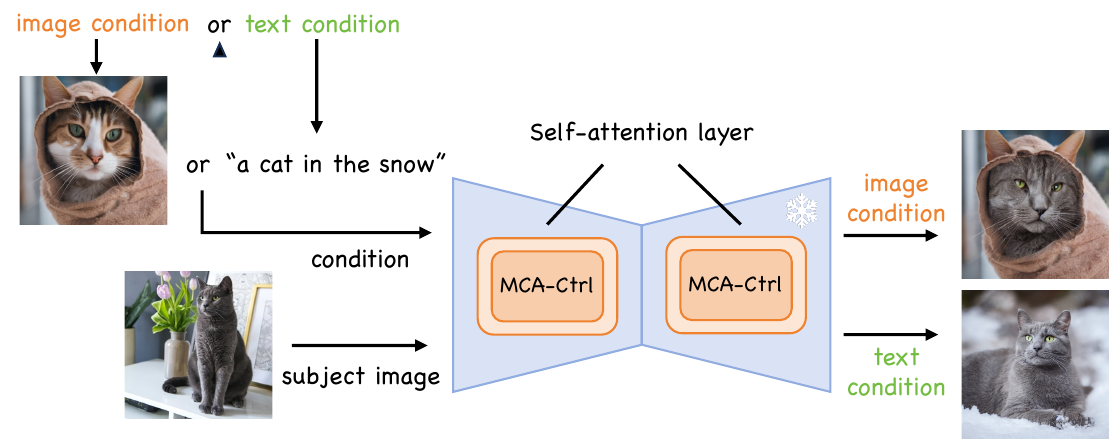}}
  \vspace{-0.2cm}
  \caption{
  The pipeline of MCA-Ctrl.
  }
  \label{intro0}
\vspace{-0.6cm}
\end{figure}

\begin{figure*}[h]
  \vspace{-0.2cm}
  \centerline{\includegraphics[width=1\linewidth]{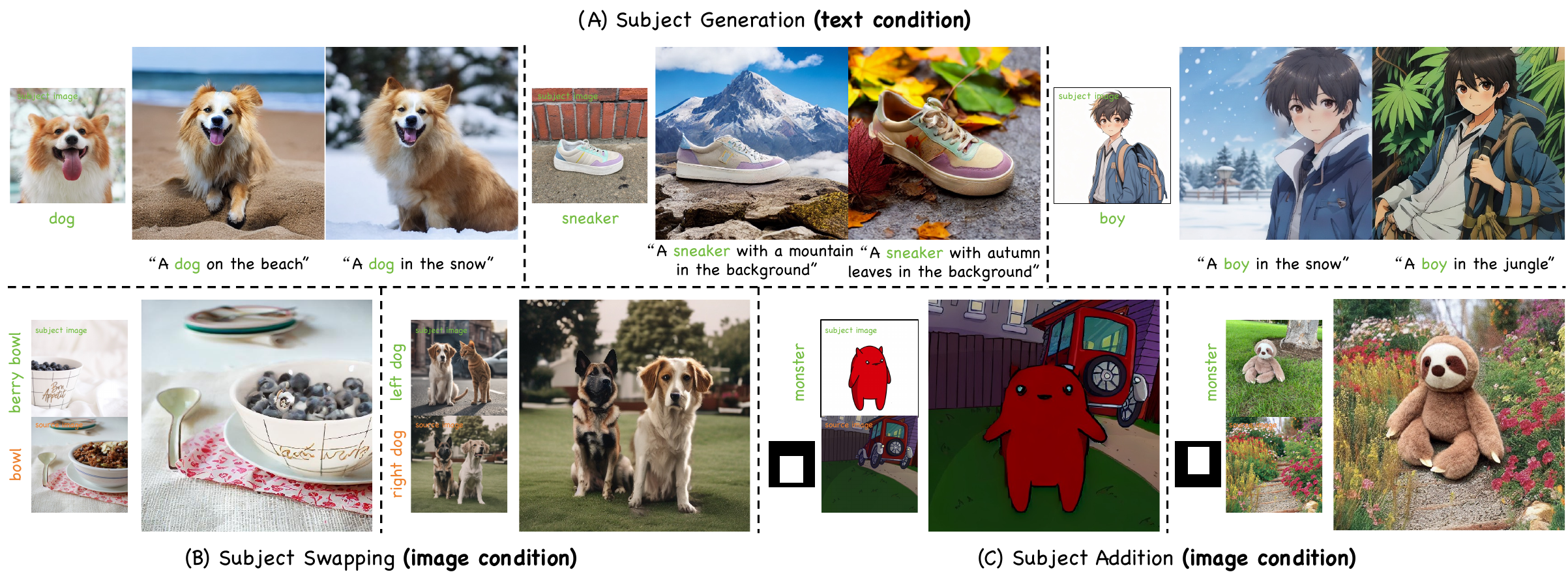}}
  \vspace{-0.2cm}
  \caption{
  Customized results from MCA-Ctrl.
  Without any fine-tuning or training, MCA-Ctrl can be used for text-driven subject image generation and image-driven subject image editing. Our method achieves high-quality customization across animals, people, and objects, preserving the distinctive features of specified subjects and meeting users’ specific requirements.
  }
  \label{intro}
\vspace{-0.5cm}
\end{figure*}

Image customization involves maintaining the identity and essence of a subject from a reference image while creating new representations under text or visual conditions.  Traditionally, this has involved inverting the visual representation of the subject into a textual latent space and reconstructing new subject images through placeholders \cite{ruiz2023dreambooth, gal2022image}. However, this process often requires extensive fine-tuning or costly optimization for each subject. To address these challenges, certain approaches, such as IP-Adapter \cite{ye2023ip} and BLIP-Diffusion \cite{li2024blip}, have been developed to reduce training costs and enhance zero-shot performance by training a multimodal encoder and an alignment projection layer between image and text representations. BLIP-Diffusion \cite{li2024blip} incorporates the transformed image representation into the prompt to guide image generation and editing. The series of works on IP-Adapter \cite{ye2023ip} treats the image representation as another form of prompt, employing the same cross-attention mechanism with text to introduce consistency.

However, whether subject representation is derived through inversion or a multimodal encoder, several limitations remain:
(1) Lower controllability, primarily text-driven. Some works \cite{ruiz2023dreambooth, gal2022image, chen2024subject, ye2023ip, huang2024realcustom} are driven solely by text, which introduce uncertainties in the background, layout, and other elements. Some recent studies \cite{li2024tuning,gu2024photoswap} suggest using image condition to enhance control over background and custom regions. However, these approaches are often limited to single applications, focusing solely on either swapping or addition, thus restricting their applicability.
(2) Subject leakage or confusion in complex visual conditions.  
We consider complex visual scenes to include object interactions, occlusions, multiple objects, and similarities between foreground and background. In these cases, inaccuracies in high-response regions during model generation will lead to subject leakage and confusion.
(3) Poor background consistency under image conditions. 
(4) High fine-tuning costs for inversion-based approaches and lower subject consistency for adapter-based methods. 
Therefore, as shown in Figure~\ref{intro0}, this paper seeks to explore a customization method \emph{compatible with both text and image conditions}, \emph{low computational costs}, and \emph{high quality}.

To achieve this goal, this paper introduces \textbf{\emph{Multi-party Collaborative Attention Control (MCA-Ctrl)}}, a tuning-free framework that enables controllable image customization under text or image conditions. 
Specifically, as shown in Figure~\ref{intro}, MCA-Ctrl can perform three types of tasks: \textbf{\emph{subject generation}}, \textbf{\emph{subject swapping}}, and \textbf{\emph{subjet addition}}. The generation task is text-driven, while the swapping and addition tasks are image-driven.
Built upon Stable Diffusion, MCA-Ctrl manipulates three flexible parallel diffusion processes within the self-attention layers to control the generation of the target image. These three diffusion processes are the subject diffusion process, the target image diffusion process, and the condition diffusion process, with the latter operating differently based on the form of the condition (text or image).
Two distinct feature interaction operations within the self-attention layers are included: Self-Attention Global Injection (SAGI) and Self-Attention Local Query (SALQ). SALQ initiates from the target image, querying key information from the subject and conditional information. SAGI starts from the subject and conditional information, injecting the necessary visual features into the target image generation process. The combination of these two operations allows the model to maintain high consistency with both the subject and conditional information without requiring fine-tuning.
To tackle subject leakage and confusion in complex visual scenarios, we introduce a Subject Localization Module (SLM) that processes multi-modal instructions. This module refines the model's high-response regions, improving MCA-Ctrl's image generation quality.

Our main contributions are as follows:
\begin{itemize}[leftmargin=*]
\item We introduce MCA-Ctrl, a tuning-free method that achieves high-quality image customization under both text and image conditions, outperforming previous approaches in quantitative metrics and human evaluations.
\item We propose two complementary attention control strategies that enable the generated images to maintain high consistency with both the target subject and the conditional information simultaneously.
\item We present a Subject Localization Module (SLM) that corrects the high-response regions of the model in complex visual scenarios, reducing artifacts caused by feature confusion.
\end{itemize}

\section{Related Works}
\subsection{Image Editing with Diffusion Models}
Recently, the text-to-image latent diffusion models proposed enable the most advanced performance in image generation \cite{rombach2022high}.
These models are trained on large-scale image-text pairs datasets and can generate images guided by open-domain text descriptions.

Given an image-text pair $I_s$ and $P$, the latent diffusion model first converts $I_s$ into a feature $z$ in the latent space through an autoencoder and then, as shown in Equ.\eqref{eq_rw:1}, Gaussian noise is progressively added to $z_0$ through a predefined Markov chain, where $\beta_t$ represents the scheduler.
By converting with $\alpha_t=\prod^{t}_{s=1}(1-\beta_s)$, we can use Equ.\eqref{eq_rw:2} to transform $z_0$ to $z_t$ at any time.
\begin{equation}
\label{eq_rw:1}
    q(z_t|z_{t-1}) = \mathcal{N}(z_t;\sqrt{1-\beta_t}z_{t-1},\beta_t\mathbf{I})
\end{equation}
\begin{equation}
\label{eq_rw:2}
    q(z_t|z_0) = \mathcal{N}(z_t;\sqrt{\alpha_t}z_0;(1-\alpha_t)\mathbf{I})
\end{equation}
Finally, the $z_t$ is transformed into a high-resolution image $I_t$ by optimizing the following objectives:
\begin{equation}
    \mathcal{L}(\theta)=\mathbb{E}_{t\sim\mathcal{U}(1,T),\epsilon_t\sim\mathcal{N}(0,\mathbf{I})}||\epsilon_t-\epsilon_\theta(z_t,t,P)||^2
\end{equation}
$\epsilon_\theta$ generally refers to a network with UNet architectures that interact with text prompt $P$ through cross-attention mechanisms at different resolutions.
In inference, random noise is selected from the Gaussian distribution $z_T \sim \mathcal{N}(0,\mathbf{I})$, and the corresponding image is generated under the guidance of the given text description.
Based on the text-to-image models, text-driven image editing has been proposed.
These works can be roughly divided into two categories. 
One category, such as InstructPix2Pix \cite{brooks2023instructpix2pix}, mainly constructs instruction-based image pair datasets $(I_s, I_t, P)$ to train latent diffusion models for editing purposes, where $I_t$ is the ideal editing result of $I_s$ under the guidance of $P$. 
The second type is to achieve image editing by controlling cross-attention or self-attention, such as Prompt-to-Prompt \cite{hertz2022prompt}, MasaCtrl \cite{cao2023masactrl} and so on.

When editing the real image $R$, we need to invert the image into the latent space to obtain the $z_T$ corresponding to $R$ \cite{song2020denoising}, and then repeat the denoising process for more detailed image editing.

\subsection{Image Customization}
As image generation models advance, the demand for customization has grown. Customization involves incorporating user-provided conditions, like images or text, into generated outputs. 
Methods such as Textual Inversion \cite{gal2022image} and Dreambooth \cite{ruiz2023dreambooth} align the visual features of user-provided images with specific text placeholders to create custom content. However, these methods require extensive fine-tuning for each subject and offer limited control over layout and background.
BLIP-Diffusion \cite{li2024blip} and IP-Adapter \cite{ye2023ip} train a projection layer using large image-text datasets to align text and image features, enabling some zero-shot generation capabilities in the trained model. However, this still involves significant storage and training costs.

Prompt-to-Prompt \cite{hertz2022prompt} and MasaCtrl \cite{cao2023masactrl} highlight the rich semantic information embedded in cross-attention and self-attention layers, leading to new methods \cite{li2023dreamedit, 2024FreeCustom, li2024tuning, gu2024photoswap} for incorporating custom information through attention control.
Some works, like TIGIC \cite{li2024tuning} and PHOTOSWAP \cite{gu2024photoswap}, use background-conditioned images for more complete customization. However, these methods often address single tasks, such as swapping, generation, or addition, and may struggle with subject confusion and leakage in complex visual conditions, limiting their applicability.
This paper introduces a flexible multi-party collaborative control mechanism that handles all three customization tasks. Additionally, we propose a subject localization module to help the model more accurately recognize subjects in complex visual conditions, resulting in high-quality customized outputs.
\section{Method}
\begin{figure*}[t]
\vspace{-0.2cm}
  \centerline{\includegraphics[width=1\linewidth]{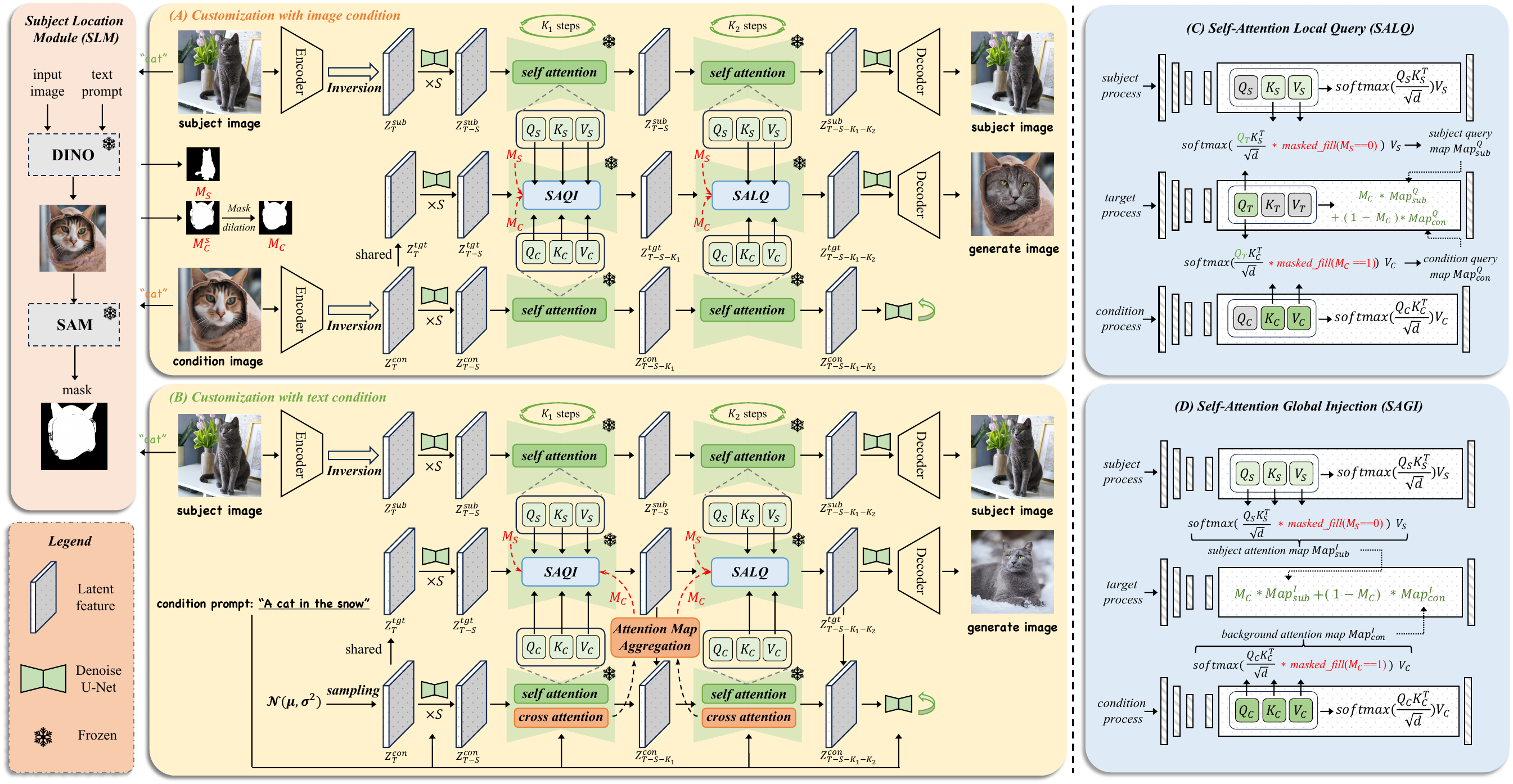}}
  \vspace{-0.2cm}
  \caption{
  \textbf{Overview of the proposed MCA-Ctrl.}
  Our method customizes images through self-attention cooperative control across three parallel diffusion processes, eliminating the need for fine-tuning. Figures (A) and (B) illustrate the inference pipeline of MCA-Ctrl under image and text conditions, while (C) and (D) show details of self-attention local query and self-attention global injection.
  }
  \label{arch}
\vspace{-0.5cm}
\end{figure*}
We propose \emph{Multi-party Collaborative Attention Control (MCA-Ctrl)}, a method that uses the knowledge inside the diffusion model for general image customization without fine-tuning.
Its core idea is to combine the semantic information of the condition image or text prompt with the content in the subject image for a novel rendition of a specific subject.
Specifically, we capture the visual appearance representation of a particular subject while preserving the spatial layout of the condition through self-attention injection and query in three parallel diffusion processes.
This task is highly challenging, and most existing customization models often require extremely costly training \cite{ruiz2023dreambooth, gal2022image, li2024blip, ye2023ip, huang2024realcustom}.

\textbf{Overall Pipeline.}
The overall pipeline for editing and generating by MCA-Ctrl is shown in Figure \ref{arch}.
MCA-Ctrl includes three diffusion processes: subject diffusion process $\mathcal{B}_{sub}$, condition diffusion process $\mathcal{B}_{con}$, and target diffusion process $\mathcal{B}_{tgt}$.
$\mathcal{B}_{sub}$ receives the real subject image $I_{sub}$ and generates the diffusion initial feature $Z^{sub}_{T}$ through a DDIM inversion \cite{song2020denoising}.
$\mathcal{B}_{con}$ receives the real source image $I_{con}$ or the text prompt $P_T$. As shown in Figure~\ref{arch} (A) and (B), for $I_{con}$, we get $Z^{con}_{T}$ the same as $\mathcal{B}_{sub}$; for $P_T$, we generate a random Gaussian distribution as $Z^{con}_{T}$.
$\mathcal{B}_{tgt}$ is a generation process that shares $Z^{con}_{T}$ with a potential spatial layout as an initial feature to generate a target image $I_T$. 
At each diffusion step, we selectively perform the following operations:
1) Inject the foreground self-attention map and background self-attention map of $\mathcal{B}_{sub}$ and $\mathcal{B}_{con}$ into $\mathcal{B}_{tgt}$, called Self-Attention Global Injection (SAGI).
2) $\mathcal{B}_{tgt}$ queries the subject appearance and background content from $\mathcal{B}_{sub}$ and $\mathcal{B}_{con}$, called Self-Attention Local Query (SALQ).
The details of SAGI and SALQ are in Section \ref{SAGI} and \ref{SALQ}.

\textbf{Subject Location Module.}
To prevent query confusion and subject feature artifacts in complex visual scenes with multiple similar objects, we introduce a Subject Location Module (SLM) to locate user-specified objects precisely. 
The SLM consists of an object detection model, DINO \cite{liu2023grounding}, and a segmentation model, SAM \cite{kirillov2023segment}. 
It processes multimodal information, such as a subject image $I_{sub}$ paired with textual prompts $P_{sub}$ and source images $I_{con}$ paired with text descriptions $P_{con}$ of regions to be edited.
After localization and segmentation, the SLM outputs a binary subject image layer $M^s_C$ and an editable image layer $M_S$. 
To ensure the edited region has sufficient space to blend with the background and avoid rigid transitions, we dilate $M^s_C$ to $M_C$ using a dilation kernel $m$ with a size of $3\times3$.

\subsection{Self-Attention Local Query (SALQ)}
\label{SALQ}
From the perspective of the task, our goal is to extract the appearance features of the subject from the subject image $I_{sub}$ and query the background content and semantic layout from the condition $I_{con}$ or $P_{T}$. 
By sharing the initial features of $\mathcal{B}_{con}$, the target image can basically form a spatial layout similar to $I_{con}$. Therefore, we focus on content queries from the condition.
Inspired by MasaCtrl \cite{cao2023masactrl}, the key feature ${K}$ and value feature ${V}$ of the self-attention layer can reflect the potential content representation of the image. 
Therefore, as shown in Figure~\ref{arch} (C), at the denoising step $t$ and layer $l$, $\mathcal{B}_{tgt}$ queries the foreground and background content from $\mathcal{B}_{sub}$ and $\mathcal{B}_{con}$ through the query feature $Q_{T,t,l}$ of the self-attention layer. 

Through Equ~\eqref{eq:1}, we obtain the attention matrices $\mathcal{A}_{T,C,t,l}, \mathcal{A}_{T,S,t,l}$ of the target image to the global regions of the condition and subject image.
To limit the query region and avoid confusion, we use $M_C$ and $M_S$ to mask the attention matrices locally, that is, to query foreground content only in the subject image and background content only in the condition.
Then, according to Equ~\eqref{eq:2} and \eqref{eq:3}, we can obtain the queried foreground and background content features.
Finally, we fused these two types of features through Equ~\eqref{eq:4}.
This operation serves two purposes:
1) $M_C$ is employed to constrain the editable image region and ensure the layout consistency with the condition again;
2) Simultaneously query the foreground and background content, realizing the replacement of specific object's appearances and enhancing the alignment of background content with the condition.
$\mathcal{MF}$ stands for mask fill.
\begin{equation}
\label{eq:1}
    \mathcal{A}_{T,S,t,l} = \frac{Q_{T,t,l}K_{S,t,l}^{T}}{\sqrt{d}}, 
    \mathcal{A}_{T,C,t,l} = \frac{Q_{T,t,l}K_{C,t,l}^{T}}{\sqrt{d}}
\end{equation}
\begin{equation}
\label{eq:2}
\mathcal{F}^Q_{T,S,t,l}=softmax(\mathcal{A}_{T,S,t,l}*\mathcal{MF}(M_S=0))V_{S,t,l}
\end{equation}
\begin{equation}
\label{eq:3}
\mathcal{F}^Q_{T,C,t,l}=softmax(\mathcal{A}_{T,C,t,l}*\mathcal{MF}(M_C=1))V_{C,t,l}
\end{equation}
\begin{equation}
\label{eq:4}
\mathcal{F}_{T,t,l}^{*} = M_C*\mathcal{F}^Q_{T,C,t,l} + (1-M_C)*\mathcal{F}^Q_{T,S,t,l}
\end{equation}

Unlike \cite{cao2023masactrl}, we need the layout of the target image to follow the condition as closely as possible, so we recommend performing SALQ starting with the U-Net decoder in the early step.

\subsection{Self-Attention Global Injection (SAGI)}
\label{SAGI}
After SALQ, we find that there are often two problems in generated images: 1) lack of authenticity in various details and 2) slight confusion with original features during the query process.
We believe this is because the query process is essentially a local fusion of original and query features, inevitably leading to feature crossing and confusion.
Therefore, we propose a global attention hybrid injection to enhance detail authenticity and content consistency of foreground and background.

As shown in Figure~\ref{arch} (D), we first compute the attention matrices $\mathcal{A}_{C,t,l}$ and $\mathcal{A}_{S,t,l}$ for the condition and subject image according to Equ~\eqref{eq:4}.
Unlike SALQ, $\mathcal{A}$ here is the original attention matrix in the reconstruction of $\mathcal{B}_{con}$ and $\mathcal{B}_{sub}$, including the mutual attention of all pixels in the image.
Based on our goal, we also use $M_C$ and $M_S$ to filter $\mathcal{A}_{C,t,l}$ and $\mathcal{A}_{S,t,l}$ locally to focus on  background and subject content. According to Equ~\eqref{eq:5} and \eqref{eq:6}, we can get the subject features and background features filtered by attention.
Note that $\mathcal{F}^I_{S,t,l}$ and $\mathcal{F}^I_{C,t,l}$ here does not interact with the foreground content of the target process.
We use Equ~\eqref{eq:7} to inject the subject features and background features into the target image’s diffusion process. By reconstructing the current feature output through replacement, we directly enhance foreground/background details while reducing feature confusion.
\begin{equation}
\label{eq:4}
    \mathcal{A}_{S,t,l} = \frac{Q_{S,t,l}K_{S,t,l}^{T}}{\sqrt{d}}, 
    \mathcal{A}_{C,t,l} = \frac{Q_{C, t,l}K_{C,t,l}^{T}}{\sqrt{d}}
\end{equation}
\begin{equation}
\label{eq:5}
\mathcal{F}^I_{S,t,l}=softmax(\mathcal{A}_{S,t,l}*\mathcal{MF}(M_S=0))V_{S,t,l}
\end{equation}
\begin{equation}
\label{eq:6}
\mathcal{F}^I_{C,t,l}=softmax(\mathcal{A}_{C,t,l}*\mathcal{MF}(M_C=1))V_{C,t,l}
\end{equation}
\begin{equation}
\label{eq:7}
\mathcal{F}_{T,t,l}^{*} = M_C*\mathcal{F}^I_{C,t,l} + (1-M_C)*\mathcal{F}^I_{S,t,l}
\end{equation}

However, it should be noted that $\mathcal{F}^I_{S,t,l}$ not only contains the content appearance but also the spatial layout information of the subject in $I_{sub}$.
Therefore, the location of SAGI needs to vary depending on the task.
In subject editing, we want the subject image to inject content features without layout structure information into the target process, without destroying the spatial layout guided by the initial features $Z_{T}^{tgt}$ and mask $M_C$.
Therefore, we recommend performing SAGI in the early denoising step when the reconstructed composition of the condition and subject images has yet to generate mature spatial information.
When doing subject generation, we want the subject content to be preserved completely, although some layout information is introduced. Therefore, we recommend continuously performing SAGI until later denoising steps.

\begin{algorithm}[t]
\begin{algorithmic}[1] 
\small
\REQUIRE  A source text-image pair ($I_{con}$,$P_{con}$), a subject text-image pair ($I_{sub}$,$P_{sub}$);
\ENSURE a generate image $I_{T}$.
\STATE $M_S, M_C = SLM((I_{con},P_{con}), (I_{sub},P_{sub}))$
\STATE $\{Z_{T}^{con}, Z_{T-1}^{con}, ...,Z_{0}^{con}\} = Inversion(I_{con})$
\STATE $\{Z_{T}^{sub}, Z_{T-1}^{sub}, ...,Z_{0}^{sub}\} = Inversion(I_{sub})$
\STATE $Z_{T}^{tgt}\leftarrow Z_{T}^{con}$
\FOR {$t=T, T-1, ..., 1$}
\STATE $\{Q_S, K_S, V_S\}\leftarrow \epsilon_\theta(Z_{t}^{sub}, t)$
\STATE $\{Q_C, K_C, V_C\}\leftarrow \epsilon_\theta(Z_{t}^{con}, t)$
\STATE $\{Q_T, K_T, V_T\}, \mathcal{F}\leftarrow \epsilon_\theta(Z_{t}^{tgt}, t)$
\STATE $\mathcal{F}^{*}\leftarrow$ \textbf{EDIT}($\{Q_T, K_T, V_T\}, \{Q_S, K_S, V_S\}, \{Q_C, K_C$, $V_C\}$, $M_S$, $M_C$)
\STATE $\epsilon \leftarrow \epsilon_\theta(Z_{t}^{tgt}, t, \mathcal{F}^{*})$
\STATE $Z_{t-1}^{tgt}\leftarrow Sample(Z_{t}^{tgt}, \epsilon)$
\ENDFOR
\RETURN $Z_{0}^{con}, Z_{0}^{sub}, Z_{0}^{tgt}$
\caption{The procedure of MCA-Ctrl for customization with image condition}
\label{alg:inference}
\end{algorithmic}
\end{algorithm}
\subsection{Inference of MCA-Ctrl}
The algorithm flow of image customization with image condition is shown in Algorithm \ref{alg:inference}. 
Assuming that the start and end steps of SAGI and SALQ are $S_{GI}$, $E_{GI}$, $S_{LQ}$, $E_{LQ}$, and the start layers are $Layer_{GI}$ and $Layer_{LQ}$, and the execution intervals of SAGI and SALQ do not cross.The \textbf{EDIT} function of Algorithm \ref{alg:inference} at denoising step $t$ and layer $l$ is as follows:
\begin{equation}
\label{eq:0}
    \textbf{EDIT}:=
    \left\{  
    	\begin{aligned}
    	&SAGI,\mbox{if}\ S_{GI} \textless t \textless E_{GI}\ \mbox{and}\ l \textgreater Layer_{GI} \\
            &SALQ,\ \mbox{if}\ S_{LQ} \textless t \textless E_{LQ}\ \mbox{and}\ l \textgreater Layer_{LQ}\\
            &Self\mbox{-}Attention(\{Q_T, K_T, V_T\}),\; \mbox{otherwise}
    	\end{aligned}
    \right.
\end{equation}
\emph{Self-Attention} represents the standard self-attention operation\cite{vaswani2017attention}.
If the condition is text prompt, the acquisition of $M_C$ is changed to extract from the cross attention of the corresponding step in $\mathcal{B}_{con}$ as shown in Figure~\ref{arch} (B).
Notably, although we present the inference of MCA-Ctrl as three parallel diffusion processes, \emph{this does not incur any additional computational cost}. In the code implementation, these three parallel diffusion processes are handled as a single inference run with a batch size of 3.

\section{Experiment}
\begin{figure*}
\vspace{-0.3cm}
  \centerline{\includegraphics[width=1\linewidth]{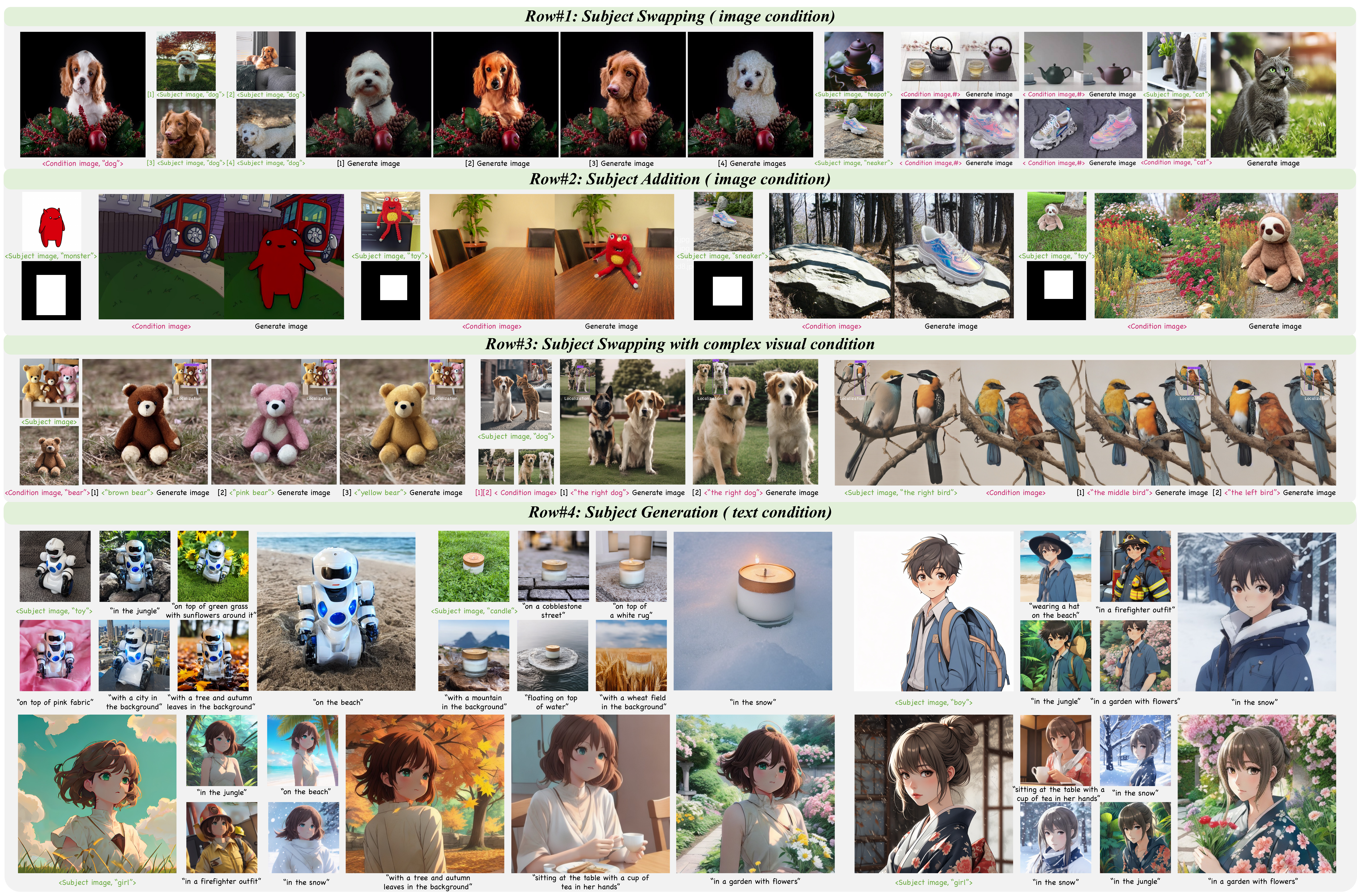}}
  \vspace{-0.2cm}
  \caption{
  Qualitative result of MCA-Ctrl.
  }
  \label{fig1}
\vspace{-0.5cm}
\end{figure*}
\subsection{Experimental Settings}
\label{setting2}
\textbf{Dataset.}
We utilize DreamBench \cite{ruiz2023dreambooth} as the subject dataset, which consists of 30 subjects such as plush animals, dogs, cats, clocks, and robots. Then, we use DreamEditBench \cite{li2023dreamedit} as the condition image dataset, providing ten editable real images for each subject in DreamBench. For subject generation, we employ 25 prompt templates from DreamBench to generate four images per prompt for model robustness assessment.

\textbf{Metrics.}
We evaluate the images using three types of metrics: DINO \cite{caron2021emerging} and CLIP-I \cite{radford2021learning} to assess image-to-image similarity, CLIP-T to evaluate image-to-text alignment, and ImageReward \cite{xu2024imagereward} to measure image aesthetic quality.
Additionally, in subject swapping and addition tasks, we further divide DINO and CLIP-I into DINO$_{sub}$, DINO$_{back}$, CLIP-I$_{sub}$, and CLIP-I$_{back}$, representing the consistency of the subject and background.

\textbf{Setup.}
Our method utilizes the latest stable text-to-image diffusion model \cite{rombach2022high} with checkpoint v1.5. We employ DDIM deterministic inversion \cite{song2020denoising} for real image editing, converting images into initial noise maps. During sampling, we conduct 50 denoising steps of DDIM sampling with classifier-free guidance \cite{ho2022classifier, yang2025hsrdiff} set to 7.5.
Unless specified, SAGI is executed first, followed immediately by SALQ with no intermediate steps, meaning $S_{LQ} = E_{GI}$. Additionally, in all experimental validations, SAGI consistently performs better across all layers of the UNet, making $Layer_{GI} = 16$ the default setting in our paper.
In summary, our experiments focus on tuning four parameters: $S_{GI}$, $E_{GI}$, $Layer_{LQ}$, and $E_{LQ}$. These parameters can be adjusted for different classes to ensure more consistent editing and generation.
For ``\textbf{Ours} (Uniform)" in Table~\ref{table1}, we use the settings $S_{GI}=0$, $E_{GI}=20$, $Layer_{LQ}=8$, and $E_{LQ}=48$. For ``\textbf{Ours} (Uniform)" in Tables~\ref{table2} and \ref{table3}, we set $S_{GI}=0$, $E_{GI}=35$, $Layer_{LQ}=0$, and $E_{LQ}=48$.

\subsection{Main Results}
\textbf{Main qualitative results.}
Figure~\ref{fig1} shows the qualitative editing and generated results of MCA-Ctrl.
\begin{figure*}[h]
\vspace{-0.3cm}
  \centerline{\includegraphics[width=1\linewidth]{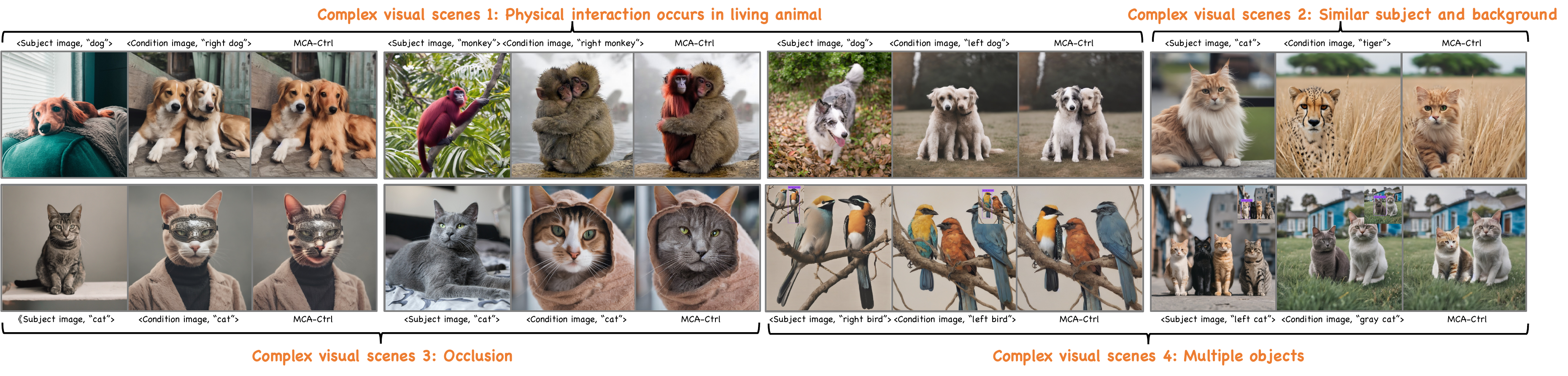}}
  \vspace{-0.3cm}
  \caption{
  Editing results of MCA-Ctrl in complex visual condition.
  }
  \label{fig2}
\end{figure*}
The first three rows primarily showcase subject editing performance, including subject swapping, subject addition, and subject swapping in complex visual scenes, demonstrating the high consistency and realism of MCA-Ctrl in both subject and background customization.
Row$\#$4 illustrates MCA-Ctrl's zero-shot customization generation capabilities, achieving high-quality, consistent, and novel reproductions across objects, animals, and people.
To further validate MCA-Ctrl's editing capabilities in complex visual scenes, we categorize such scenarios into four types: \emph{Physical interactions between subjects}, \emph{Similar subject and background}, \emph{Occlusion}, and \emph{Multiple objects}. Figure~\ref{fig2} provides examples for each. The results show that MCA-Ctrl accurately captures the appearance of different subjects in complex scenes based on user instructions, enabling high-quality edits of specified subjects within multi-object conditions.
Our model is unrestricted by manually curated datasets, allowing it to capture features from any subject in the diffusion process, with strong generalization and robustness.
\begin{table}[h]
\vspace{-0.2cm}
  \caption{
  Quantitative comparisons on DreamEditBench of subject swapping. 
  \textbf{Ours} (Uniform) means that all classes are tested with uniform parameters of $S_{GI}$, $E_{GI}$, $Layer_{LQ}$ and $E_{LQ}$; \textbf{Ours} (Specified) means to customize parameters for partial classes.
  }
  \vspace{-0.2cm}
  \label{table1}
  \centering
  \resizebox{1\linewidth}{!}{
  \begin{tabular}{l|ccccc}
    \toprule
    \textbf{Methods} & \textbf{DINO$_{sub}$}\textcolor{green}{$\uparrow$} & \textbf{DINO$_{back}$}\textcolor{green}{$\uparrow$} & \textbf{CLIP-I$_{sub}$}\textcolor{green}{$\uparrow$} & \textbf{CLIP-I$_{back}$}\textcolor{green}{$\uparrow$} & \textbf{ImageReward}\textcolor{green}{$\uparrow$} \\
    \hline
    DreamBooth \cite{ruiz2023dreambooth}  & \underline{0.6400} & 0.4270 & \underline{0.8110} & 0.7360 &  -1.1713 \\
    Customized-DiffEdit \cite{couairon2022diffedit}  & 0.5100 & \textbf{0.7850} & 0.7550 & \textbf{0.8950} & 0.1375 \\
    DreamEditor(5) \cite{li2023dreamedit}  & 0.5640 & 0.6670 & 0.7700 & 0.8550 & -0.5633 \\
    $\qquad$-iteration=1  & 0.5460 & 0.6640 & 0.7630 & 0.8530 & -0.2731 \\
    BLIP-Diffusion \cite{li2024blip}  & 0.6155 & 0.6392 & 0.8009 & 0.8248 & 0.2187 \\
    PHOTOSWAP \cite{gu2024photoswap} & 0.6307 & 0.6072 & 0.7886 & 0.7977 & -0.1982 \\
    \hline
    \textbf{Ours} (Uniform) & 0.6327$\pm$0.004 & 0.6684$\pm$0.004 & 0.7794$\pm$0.003 & 0.8621$\pm$0.005 & \underline{0.2728}$\pm$0.05 \\
    \textbf{Ours} (Specified) & \textbf{0.6433}$\pm$0.005 & \underline{0.6782}$\pm$0.002 & \textbf{0.8113}$\pm$0.004 & \underline{0.8681}$\pm$0.004 & \textbf{0.3214}$\pm$0.05 \\
    \bottomrule
  \end{tabular}}
  \vspace{-0.3cm}
\end{table}

\begin{table}[ht]
\centering
\caption{Automatic Evaluation on the DreamBench of subject generation.}
\vspace{-0.2cm}
\label{table2}
\resizebox{1\linewidth}{!}{
\begin{tabular}{l|cccc}
\toprule
\textbf{Methods} & \textbf{DINO}\textcolor{green}{$\uparrow$} & \textbf{CLIP-I}\textcolor{green}{$\uparrow$} & \textbf{CLIP-T}\textcolor{green}{$\uparrow$} & \textbf{ImageReward}\textcolor{green}{$\uparrow$} \\
\hline
DreamBooth \cite{ruiz2023dreambooth} & 0.6680 & \underline{0.8430} & \textbf{0.3060} & \underline{0.3839} \\
Textual Inversion \cite{gal2022image} & 0.5690 & 0.7800 & 0.2550 & -0.9788 \\
Re-Imagen \cite{chen2022re} & 0.6000 & 0.7900 & 0.2700 & -0.1765\\
BLIP-Diffusion \cite{li2024blip} & \underline{0.6700} & 0.8250 & 0.3020 & 0.1829 \\
IP-Adapter \cite{ye2023ip} & 0.6504 & 0.8232 & 0.2651 & -0.1782 \\
FreeCustom \cite{2024FreeCustom} & 0.6660 & 0.8363 & 0.2829 & -1.1723 \\
\hline
\textbf{Ours} (Uniform) & 0.6610$\pm$0.002 & 0.8399$\pm$0.003& 0.3022$\pm$0.002 & 0.3037$\pm$0.05 \\
\textbf{Ours} (Specified) & \textbf{0.6724}$\pm$0.004 & \textbf{0.8441}$\pm$0.003 & \underline{0.3056}$\pm$0.002 & \textbf{0.4132}$\pm$0.06 \\
\bottomrule
\end{tabular}}
\vspace{-0.2cm}
\end{table}

\begin{table}[ht]
\centering
\caption{Human Evaluation on the DreamBench of subject-driven generation.}
\vspace{-0.2cm}
\label{table3}
\resizebox{1\linewidth}{!}{
\begin{tabular}{l|c|ccc|c}
\toprule
\textbf{Methods} & \textbf{Backbone} & \textbf{Subject}\textcolor{green}{$\uparrow$} & \textbf{Textual}\textcolor{green}{$\uparrow$} & \textbf{Realistic}\textcolor{green}{$\uparrow$} & \textbf{Overall}\textcolor{green}{$\uparrow$} \\
\hline
DreamBooth \cite{ruiz2023dreambooth} & SD \cite{rombach2022high} & 0.81 & 0.64  & 0.91 & 2.36 \\
Textual Inversion \cite{gal2022image} & SD \cite{rombach2022high} & 0.44 & 0.76 & 0.86 & 2.06 \\
Re-Imagen \cite{chen2022re} & Imagen \cite{saharia2022photorealistic} & 0.71 & 0.79 & 0.80 & 2.3  \\
BLIP-Diffusion \cite{li2024blip} & SD \cite{rombach2022high} & 0.85 & 0.82 & \underline{0.93} & 2.6 \\
IP-Adapter \cite{ye2023ip} & SD \cite{rombach2022high} & 0.85 & \underline{0.84} & \textbf{0.94} & \underline{2.63} \\
FreeCustom \cite{2024FreeCustom} & SD \cite{rombach2022high} & 0.87 & 0.82 & 0.81 & 2.6 \\
\hline
\textbf{Ours} (Uniform) & SD\cite{rombach2022high} & \underline{0.88} & \underline{0.84} & 0.85 & 2.57 \\
\textbf{Ours} (Specified) & SD\cite{rombach2022high} & \textbf{0.92} & \textbf{0.89} & 0.92 & \textbf{2.73} \\
\bottomrule
\end{tabular}}
\vspace{-0.3cm}
\end{table}

\textbf{Comparison.}
\begin{figure*}[t]
\vspace{-0.2cm}
  \centerline{\includegraphics[width=1\linewidth]{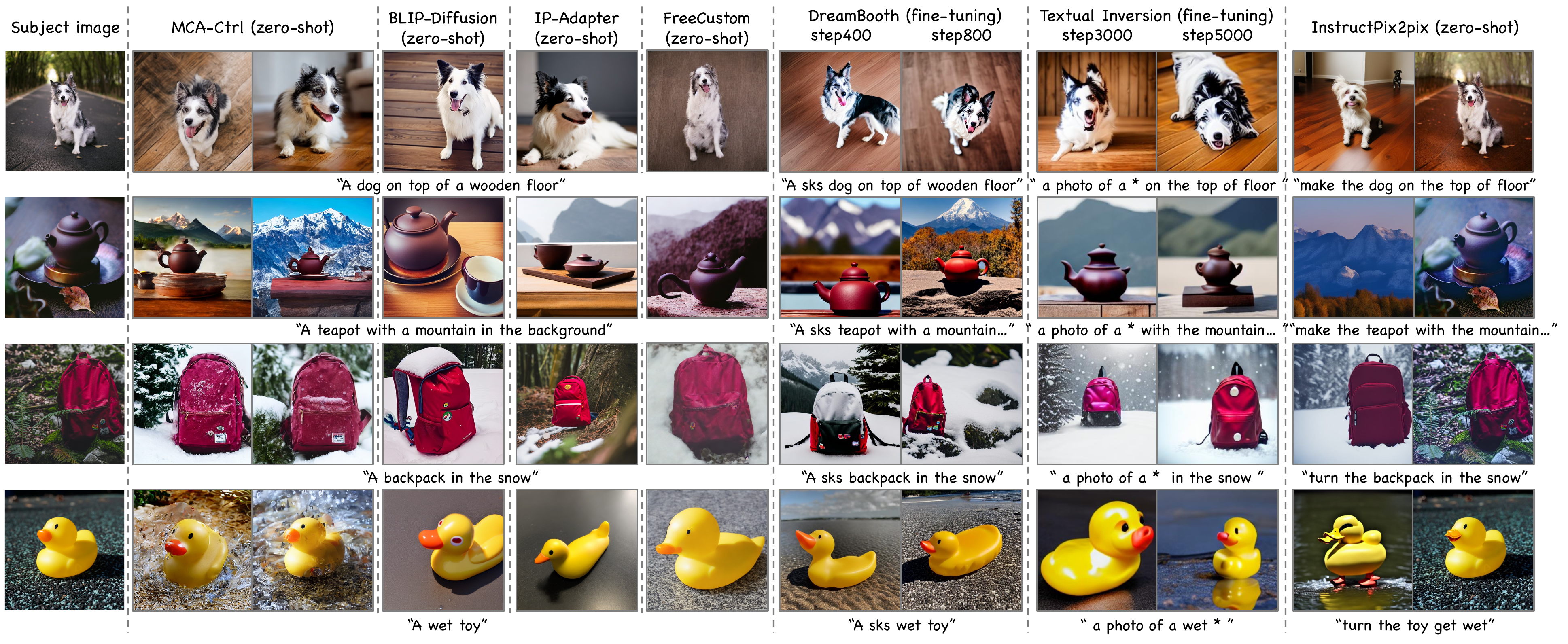}}
  \vspace{-0.3cm}
  \caption{
  Comparison with other Subject-driven Generation Models.
  }
  \label{fig3}
\vspace{-0.5cm}
\end{figure*}
\begin{figure}[h]
  \centerline{\includegraphics[width=1\linewidth]{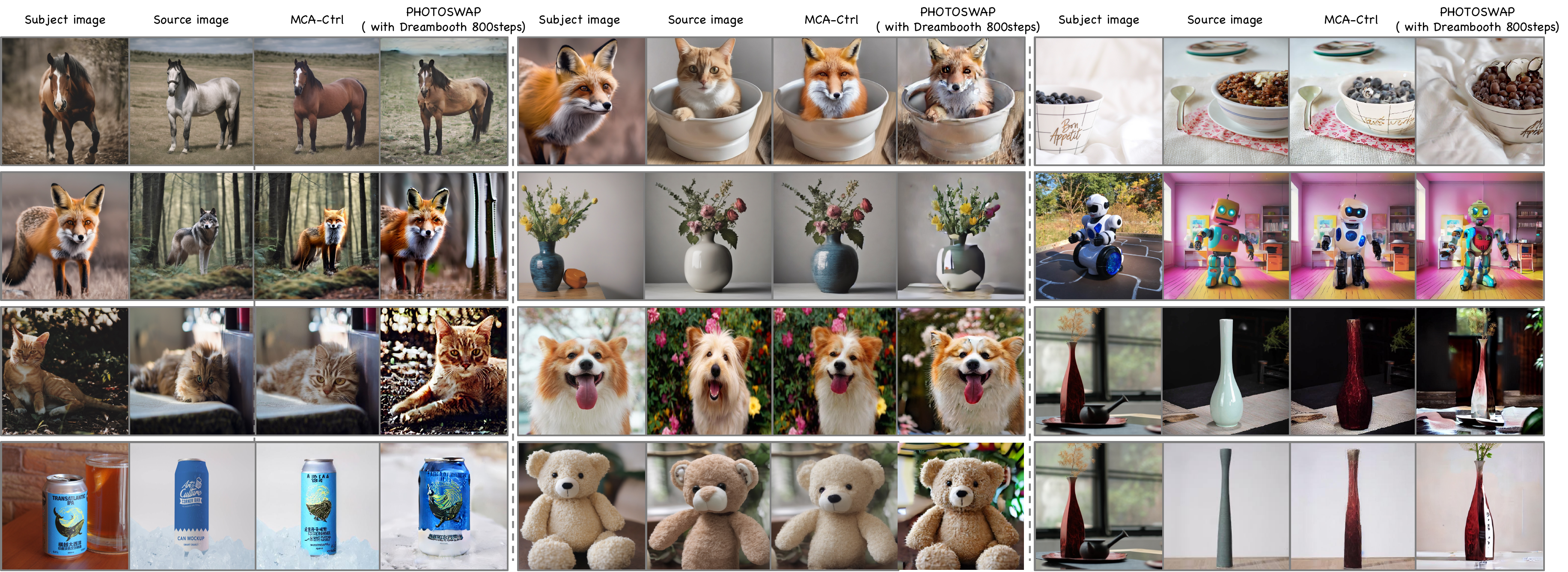}}
  \vspace{-0.2cm}
  \caption{
  Qualitative comparison between MCA-Ctrl and PHOTOSWAP on controllable subject editing.
  }
  \label{fig4}
\vspace{-0.6cm}
\end{figure}
Table~\ref{table1} presents the quantitative automatic evaluation results for the subject swapping task assessed on DreamEditBench \cite{li2023dreamedit}.
MCA-Ctrl demonstrates comparable or superior performance across all metrics relative to BLIP-Diffusion \cite{li2024blip}, DreamBooth \cite{ruiz2023dreambooth} and PHOTOSWAP \cite{gu2024photoswap}.
Specifically, with uniform parameters, MCA-Ctrl achieves slightly higher scores than BLIP-Diffusion in DINO$_{sub}$, DINO$_{back}$, CLIP-I$_{back}$, CLIP-T, and ImageReward, while recording marginally lower scores than DreamBooth in DINO$_{sub}$.
Upon adjusting parameters for some classes, MCA-Ctrl surpasses DreamBooth in DINO$_{sub}$ and CLIP-I$_{sub}$, thus indicating superior editing quality.
As shown in Figure~\ref{fig4}, as a training-free method, MCA-Ctrl outperforms PHOTOSWAP in capturing subject features while preserving the original layout and background content of the image.
Detailed scores both before and after parameter adjustment for each subejct and the specific scheme for parameter adjustment are shown in \emph{Supplementary material}.
Note that, in the reported result \textbf{Ours} (Specified), we make only subtle adjustments to the execution steps of SAGI and the execution layers of SALQ for certain classes. Overall, these adjustments are easy to implement and not time-consuming. 

\begin{figure}[h]
\vspace{-0.2cm}
  \centerline{\includegraphics[width=1\linewidth]{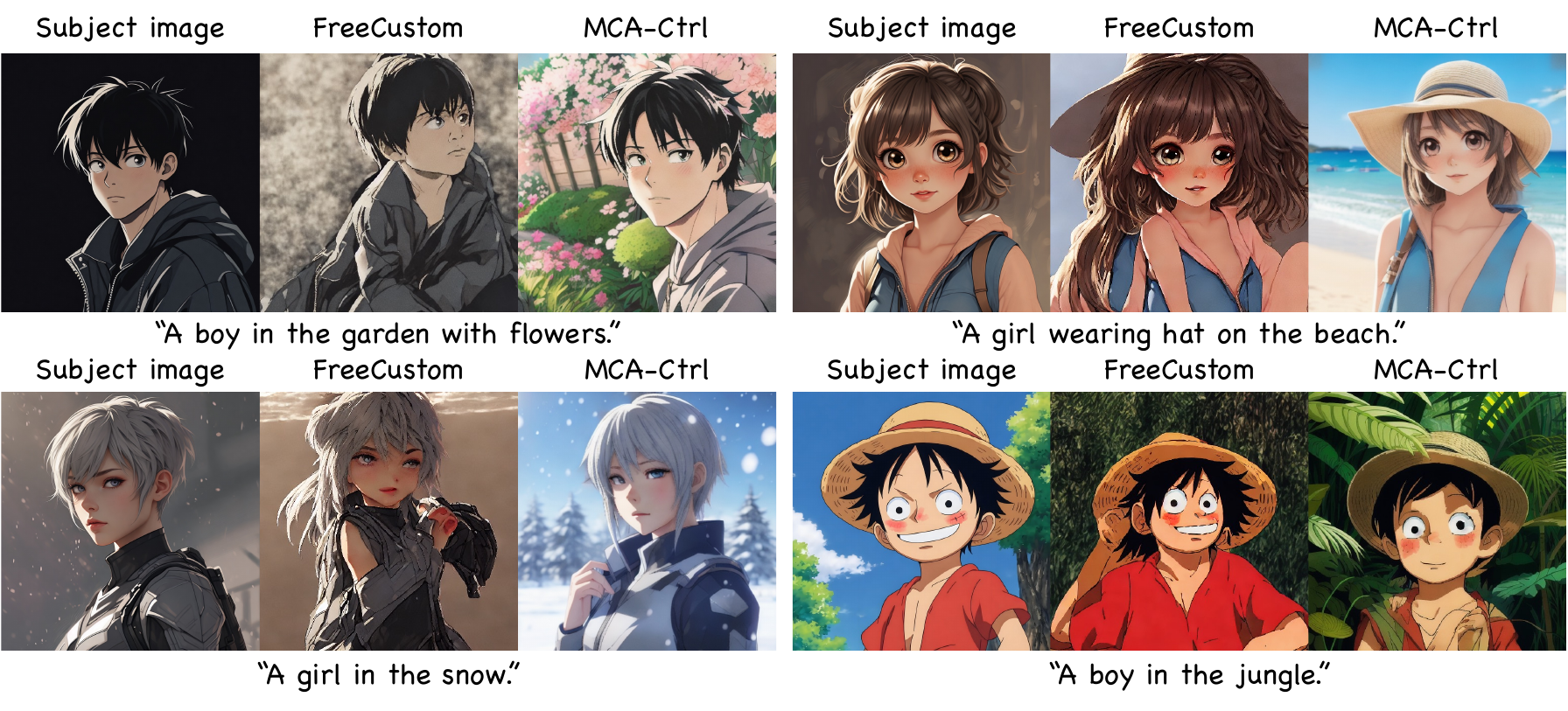}}
  \vspace{-0.2cm}
  \caption{
  Comparison between MCA-Ctrl and FreeCustom on character customization.
  }
  \label{fig5}
  \vspace{-0.2cm}
\end{figure}
Table~\ref{table2} shows automatic evaluation results for the subject generation task on DreamBench. Initially, MCA-Ctrl performs better than Text Inversion, Re-Imagen, and IP-Adapter but slightly lower than DreamBooth and BLIP-Diffusion with uniform parameters. However, MCA-Ctrl with specified parameters achieves results comparable to those of BLIP-Diffusion and DreamBooth. Furthermore, Table~\ref{table3} presents our human evaluation results on DreamBench, indicating that MCA-Ctrl demonstrates superior subject alignment and text alignment, slightly outperforming BLIP-Diffusion in overall score.
As a training-free method, maintaining consistency with high-granularity subjects like character figures is quite challenging. As shown in Figure~\ref{fig5}, FreeCustom struggles with errors in character customization, failing to accurately represent both the subject and background. In contrast, MCA-Ctrl overcomes this challenge through complementary multi-party collaborative control, achieving effective and accurate customization for character subjects.

\begin{table}
  \caption{
  Ablation results on DreamEditBench\cite{li2023dreamedit}.
  “reverse” means to reverse the execution order of SAGI and SALQ, executing SALQ before SAGI.
  }
  \vspace{-0.2cm}
  \label{table4}
  \centering
  \resizebox{1\linewidth}{!}{
  \begin{tabular}{l|ccccc}
    \toprule
    \textbf{Ablation setups} & \textbf{DINO$_{sub}$}\textcolor{green}{$\uparrow$} & \textbf{DINO$_{back}$}\textcolor{green}{$\uparrow$} & \textbf{CLIP-I$_{sub}$}\textcolor{green}{$\uparrow$} & \textbf{CLIP-I$_{back}$}\textcolor{green}{$\uparrow$} & \textbf{ImageReward}\textcolor{green}{$\uparrow$} \\
    \hline
    \textbf{Ours} (Uniform) & \textbf{0.6327} & 0.6684 & \textbf{0.7794} & 0.8621 & \textbf{0.2728} \\
    \hline
     - w/o SALQ & 0.4238$\downarrow$ & 0.7491$\uparrow$ & 0.7416$\downarrow$ & 0.8774$\uparrow$ & 0.2454$\downarrow$ \\
     - w/o SAGI & 0.5896$\downarrow$ & 0.6851$\uparrow$ & 0.7746$\downarrow$ & 0.8429$\downarrow$ & 0.2716$\downarrow$ \\
     - w/o mask dilation & 0.5611$\downarrow$ & 0.7319$\uparrow$ & 0.7671$\downarrow$ & 0.8754$\uparrow$ & 0.2671$\downarrow$ \\
     - w/o SLM & 0.4914$\downarrow$ & \textbf{0.8244}$\uparrow$ & 0.7532$\downarrow$ & \textbf{0.8999}$\uparrow$ & 0.1911$\downarrow$ \\
     - reverse & 0.4585$\downarrow$ & 0.5547$\downarrow$ & 0.7230$\downarrow$ & 0.8014$\downarrow$ & 0.1076$\downarrow$ \\
    \bottomrule
  \end{tabular}}
\vspace{-0.2cm}
\end{table}

\begin{figure}
\centerline{\includegraphics[width=1\linewidth]{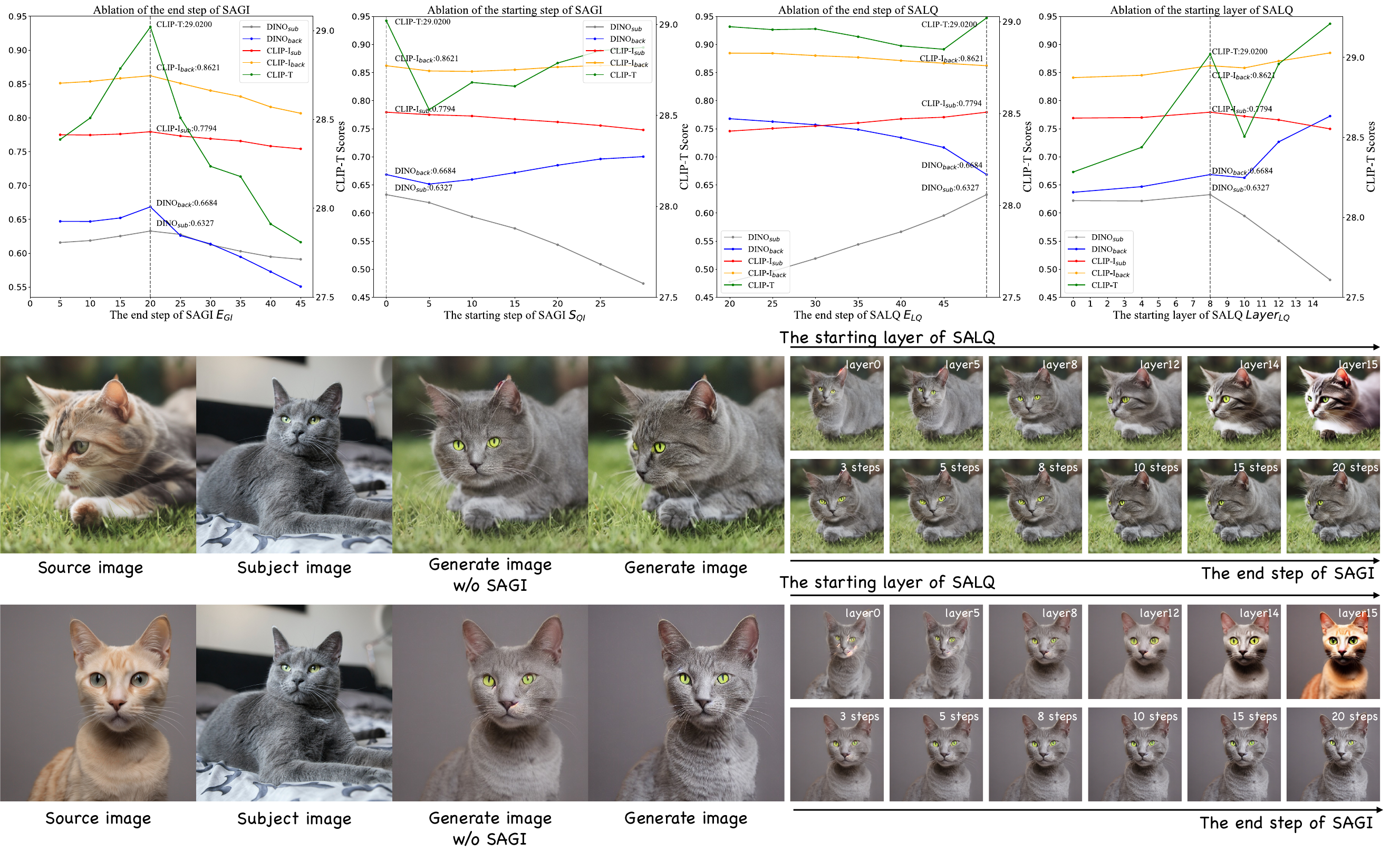}}
  \vspace{-0.2cm}
  \caption{
  \textbf{Top:} Quantitative ablation of $S_{GI}$, $E_{GI}$, $Layer_{LQ}$ and $E_{LQ}$; \textbf{Bottom:} Qualitative ablation results of SAGI and SALQ.
  Enlarged version please refer to \emph{Supplementary material}.
  }
  \label{abla1}
\vspace{-0.6cm}
\end{figure}

\textbf{Ablation Studies}
Table~\ref{table4} shows the zero-shot ablation results of MCA-Ctrl on DreamBench.
Figure~\ref{abla1} further shows quantitative and qualitative ablation of SAGI and SALQ related parameters.
Combined with the chart, we find:
\textbf{a)} SALQ is crucial. It guarantees the consistency of the generated image with the foreground appearance of the subject image, so it can significantly affect the DINO$_{sub}$ and CLIP-I$_{sub}$ scores.
\textbf{b)} SAGI can further improve the authenticity of the edited image in every detail and can correct the feature obfuscations caused by SALQ (the orange feature of the cat's mouth in Figure~\ref{abla1}), resulting in modest improvements in most metrics.
\textbf{c)} SLM can help position the specified objects when the background of the subject image or edited image is complex to improve the confusion between the foreground and background and the quality of the generated image.
\textbf{d)} The execution of SALQ from the self-attention mechanism of the encoder (0-7 layers) may cause image deformation since the layout is not yet formed. Starting from the low-resolution layer of the decoder (8-16 layers), it can inject subject features while maintaining the design of the source image. With the increase of the starting layer, the subject characteristics gradually weaken.
\textbf{e)} For the subject editing, SAGI is suitable for earlier steps, emphasizing semantic information about the foreground and background at the beginning of editing. Performing too many steps may cause the layout of the generated image foreground to be too close to the subject image.

In general, although adding certain modules may reduce consistency with the source image, qualitative and quantitative results show significant improvement in consistency with the subject image, making these trade-offs acceptable.

\textbf{Discussion}
\begin{figure}
\centerline{\includegraphics[width=1\linewidth]{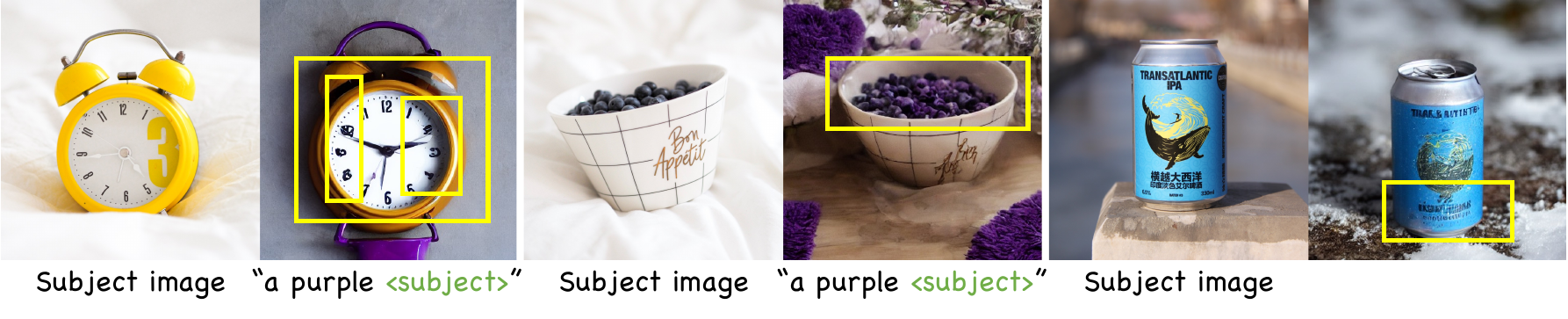}}
  \vspace{-0.2cm}
  \caption{
  Limitation of MCA-Ctrl.
  }
  \label{failure}
\vspace{-0.5cm}
\end{figure}
As shown in Figure~\ref{failure}, through extensive validation, we found that MCA-Ctrl is constrained by the base model and encounters difficulties in certain cases: (1) when the subject image contains fine-grained features, such as text; (2) when color changes are applied, there may be issues where the color change only affects the subject's local regions. Addressing these issues will be a focus of our future work.

\section{Conclusion}
This paper presents MCA-Ctrl, a tuning-free generation method for image customization. The model achieves high-quality and high-fidelity subject-driven editing and generation through coordinated attention control among three parallel diffusion processes. In addition, MCA-Ctrl solves the feature obfuscation problem in complex visual scenes by introducing a Subject Localization Module. Many experimental results show that MCA-Ctrl performs better editing and generation than most fine-tuning models.

\section{Acknowledgments}
This work is partially supported by the National Natural Science Foundation of China under Grant Number 62476264 and 62406312, the Postdoctoral Fellowship Program and China Postdoctoral Science Foundation under Grant Number BX20240385 (China National Postdoctoral Program for Innovative Talents), the Beijing Natural Science Foundation under Grant Number 4244098, and the Science Foundation of the Chinese Academy of Sciences.

{
    \small
    \bibliographystyle{ieeenat_fullname}
    \bibliography{main}
}

\appendix
\clearpage
\setcounter{page}{1}
\maketitlesupplementary

\section{Baseline Method}
We compare MCA-Ctrl to the subject-driven editing and generation methods on DreamBench \cite{ruiz2023dreambooth} and DreamEditBench \cite{li2023dreamedit} public datasets.
This section provides a brief introduction to these methods:
\begin{itemize}[leftmargin=*]
\item DreamBooth \cite{ruiz2023dreambooth}: It's a method of fine-tuning for each subject, optimizing all U-Net parameters and placeholder embedding.
\item Textual Inversion \cite{gal2022image}: This method fine-tunes each subject, optimizing the placeholder embeddings to reconstruct the subject image.
It takes 3,000 training steps to learn new concepts.
\item Re-Imagen \cite{chen2022re}: A tuning-free method that takes several images as input and then focuses on retrieval to generate new images.
\item BLIP-Diffusion \cite{li2024blip}: The model learns the multimodal subject representation step by step through the multi-modal control capability of built-in BLIP-2, achieving a certain degree of zero-shot subject-driven generation.
\item Customized-DiffEdit \cite{couairon2022diffedit}: This is a method that needs fine-tuning.
DiffEdit automatically generates the mask to be edited by contrasting predictions conditioned between the source and subject prompts.
In this paper, we follow \cite{li2023dreamedit} and replace the diffusion model in DiffEdit with the DreamBooth fine-tuned model to implement subject editing.
The generated image of this method is highly consistent with the condition image, but the foreground and connecting parts will appear stiff and have semantic incongruity.
\item DreamEditor \cite{li2023dreamedit}: This method needs fine-tuning for each class. It is implemented based on Stable Diffusion, GLIGEN, or copy-paste, and refines the target subject through iterative generation.
\item InstructPix2Pix \cite{brooks2023instructpix2pix}: A tuning-free instruction-driven editing method that takes the source image and editing instructions as input.
Although it does not explicitly express the subject, it can be a novel representation of the subject by redefining the context. We make a qualitative comparison with this method.
\item IP-Adapter \cite{ye2023ip}: A tuning-free method primarily designed for consistency-based generation.
\item FreeCustom \cite{2024FreeCustom}: A tuning-free method that leverages attention control to achieve multi-concept composition.
\item PHOTOSWAP \cite{gu2024photoswap}: A tuning-free method that enables subject swapping based on the input subject and condition images.
\item TIGIC \cite{li2024tuning}: A tuning-free method that enables subject addition based on the input subject image, condition image, and localization mask.
\end{itemize}

\section{Experimental Setting}

\subsection{Computational Efficiency}
Our three parallel diffusion processes are implemented in code by concatenating operations in the batch size dimension, i.e. each time for inference, our input shape is [3, C, H, W]. In the Self-Attention layer, we obtain the features corresponding to the subject, condition, and target images by segmenting Q, K, and V matrices and carrying out SALQ and SAGI operations. This paper describes three parallel diffusion processes to display the interaction among the subject image, condition, and target image more clearly and intuitively. \textbf{Therefore, MCA-Ctrl does not cause redundant computing resource load, and its computational efficiency is the same as that of a single execution of Stable Diffusion under the same batch size.} 

\subsection{Architecture of Subject Location Module}
As described in Section 3, the Subject Location Module consists of an object detection model Grounding DINO and a segmentation model SAM that receives a multimodal image-text pair as input and outputs a prompt-specified mask. Table \ref{SLM} lists the parameters of the Grounding DINO and SAM used in this document (All parameters that do not appear in the following table use the default parameters).
\begin{table}[h]
\begin{center}
\vspace{-0.1cm}
\caption{Specific important parameters of the model used in the Subject Location Module.}
\label{SLM}
\resizebox{1\linewidth}{!}{
\begin{tabular}{c|c|c}
\hline
 Model & parameter & value \\
\hline
\multirow{8}{*}{DINO}& backbone & swin$\_$B$\_$384$\_$22k\\
& position$\_$embedding & sine \\
& enc$\_$layers & 6 \\
& dec$\_$layers & 6 \\
& hidden$\_$dim & 6 \\
& nheads & 8 \\
& box$\_$threshold & 0.3 \\
& text$\_$threshold & 0.25 \\
 \hline
SAM& checkpoint & sam$\_$vit$\_$h$\_$4b8939.pth \\
\hline
\end{tabular}}
\end{center}
\vspace{-0.4cm}
\end{table}

\subsection{Specific Parameters of SALQ and SAGI}
As stated in Section 3.3 and Section 4.1, a total of six parameters are involved in the experiment in this paper, namely $S_{GI}$, $E_{GI}$, $S_{LQ}$, $E_{LQ}$, $Layer_{GI}$and $Layer_{LQ}$. Based on all the experimental verification, we set two default settings to make the model generation effect better: (1) SALQ is carried out continuously after SAGI operation, there is no gap between them, and the two operations do not overlap, so $E_{GI}$=$S_{LQ}$; (2) If SAGI is performed at a time step, it is performed at all layers in UNet, so $Layer_{GI}$=0. Based on the above assumptions, we mainly discussed the following four parameters: $S_ {GI} $, $E_ {GI} $, $Layer_ {LQ} $, and $E_ {LQ} $. These parameters can be adjusted for different classes to ensure more consistent editing and generation.

In Table \ref{Generation} and Table \ref{Edit}, we supplement the specific parameter settings of Our (Uniform) and Ours (Specified) models mentioned in the presentation of quantitative results for subject generation and subject swapping to help the reader reproduce the results (uniform parameter settings are used for classes not mentioned in the table).

\begin{table}[h]
\begin{center}
\caption{Specific parameters of SALQ and SAGI (Swapping).}
\label{Edit}
\resizebox{1\linewidth}{!}{
\begin{tabular}{c|c|c|c|c}
\hline
\multicolumn{5}{c}{Ours (Uniform)} \\
\hline
 Subjects & $S_{GI}$ & $Layer_{LQ}$ & $E_{GI}$ & $E_{LQ}$ \\
 \hline
 All & 0 & 8 & 20 & 48 \\
 \hline
 \multicolumn{5}{c}{Ours (Specified)} \\
 \hline
 Subjects & $S_{GI}$ & \textcolor{green}{$Layer_{LQ}$} & \textcolor{green}{$E_{GI}$} & $E_{LQ}$ \\
 \hline
 backpack & 0 & 0 & 15 &48 \\
 backpack-dog & 0 & 10 & 35 &48 \\
 berry-bowl & 0 & 10 & 17 &48 \\
 can & 0 & 8 & 10 &48 \\
 colorful-sneaker & 0 & 8 & 15 &48 \\
 dog& 0& 10 & 10 &48 \\
 dog2& 0& 10 & 25 &48 \\
 dog5& 0& 8 & 15 &48 \\
 dog6& 0& 10 & 10 &48 \\
 dog8& 0& 10 & 10 &48 \\
 duck-toy& 0& 8 & 15 &48 \\
 fancy-boot& 0& 10 & 30 &48 \\
 wolf-plushie& 0& 10 & 5 &48 \\
\hline
\end{tabular}}
\end{center}
\end{table}

\begin{table}[h]
\begin{center}
\vspace{-0.3cm}
\caption{Specific parameters of SALQ and SAGI (Generation).}
\label{Generation}
\resizebox{1\linewidth}{!}{
\begin{tabular}{c|c|c|c|c}
\hline
\multicolumn{5}{c}{Ours (Uniform)} \\
\hline
 Subjects & $S_{GI}$ & $Layer_{LQ}$ & $E_{GI}$ & $E_{LQ}$ \\
 \hline
 All & 0 & 0 & 35 & 48 \\
 \hline
 \multicolumn{5}{c}{Ours (Specified)} \\
 \hline
 Subjects & $S_{GI}$ & $Layer_{LQ}$ & \textcolor{green}{$E_{GI}$} & $E_{LQ}$ \\
 \hline
 backpack & 0 & 0 & 25 &48 \\
 backpack-dog & 0 & 0 & 30 &48 \\
 berry-bowl & 0 & 0 & 30 &48 \\
 can & 0 & 0 & 40 & 48 \\
 colorful-sneaker & 0 & 0 & 40 &48 \\
 cat& 0& 0 & 25 &48 \\
 dog& 0& 0 & 30 &48 \\
 dog2& 0& 0 & 30 &48 \\
 dog5& 0& 0 & 30 &48 \\
 dog8& 0& 0 & 30 &48 \\
 duck-toy& 0& 0 & 25 &48 \\
 fancy-boot& 0& 0 & 40 &48 \\
 wolf-plushie& 0& 0 & 25 &48 \\
\hline
\end{tabular}}
\end{center}
\vspace{-0.5cm}
\end{table}

\subsection{Analysis of $E_{GI}$}
\begin{figure*}
\centerline{\includegraphics[width=1\linewidth]{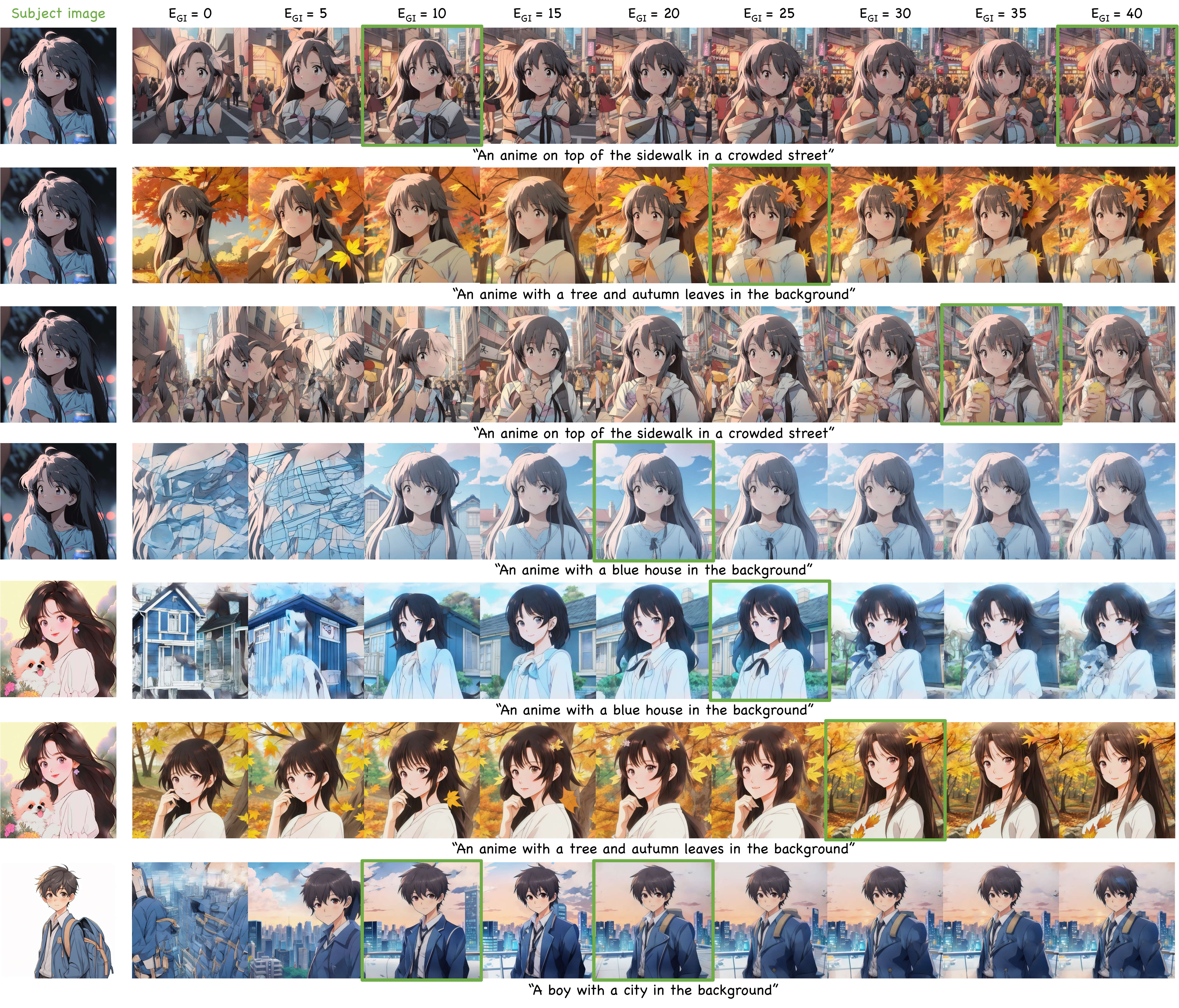}}
\vspace{-0.3cm}
  \caption{
  Analysis of $E_{GI}$.The results above are generated with a total of 50 denoising steps. Cases with green borders represent those with better performance.
  }
  \label{fig6}
  \vspace{-0.3cm}
\end{figure*}

We illustrate the impact of $E_{GI}$ on image generation in Figure \ref{fig6}. In complex scenarios, omitting SAGI can lead to challenges such as failing to localize the target and confusion in global features. As $E_{GI}$ is delayed, subject features become increasingly distinct. However, beyond a certain point (empirically around 60$\%$ of the total denoising steps for most cases), further increasing the execution steps of SAGI has a diminishing effect on image quality.

\subsection{More Visualization}
We present enlarged versions of Figures 7, 8, and 9 from the main text in Figures~\ref{fig1}, \ref{fig2}, and \ref{fig3}, respectively.

To further demonstrate the zero-shot generation capability of MCA-Ctrl, we provide additional results in Figures \ref{fig4} and \ref{fig5}. As shown, MCA-Ctrl excels at customized generation for high fine-grained objects such as animals and characters, achieving remarkable text-image and image-image consistency in the results.
\begin{figure*}[h]
  \centerline{\includegraphics[width=1\linewidth]{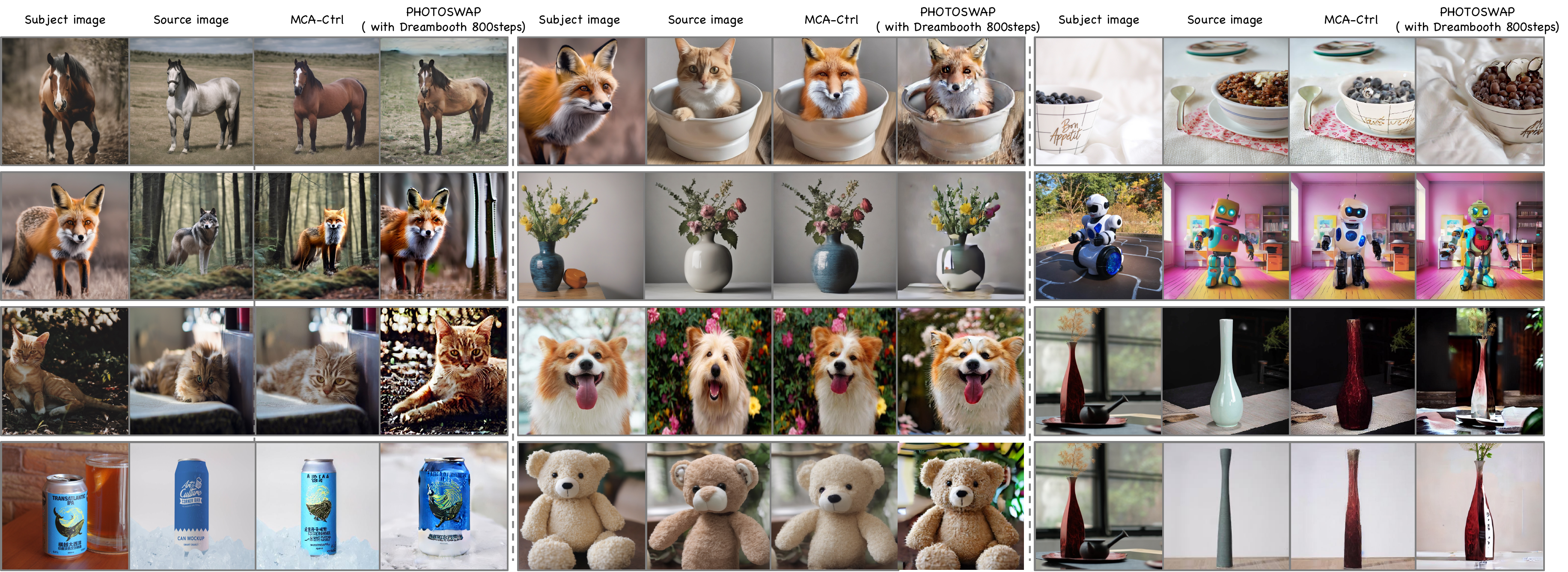}}
  \vspace{-0.1cm}
  \caption{
  Enlarged version of Figure 7 in the main text.
  }
  \label{fig1}
  \vspace{-0.2cm}
\end{figure*}

\begin{figure*}[h]
  \centerline{\includegraphics[width=0.95\linewidth]{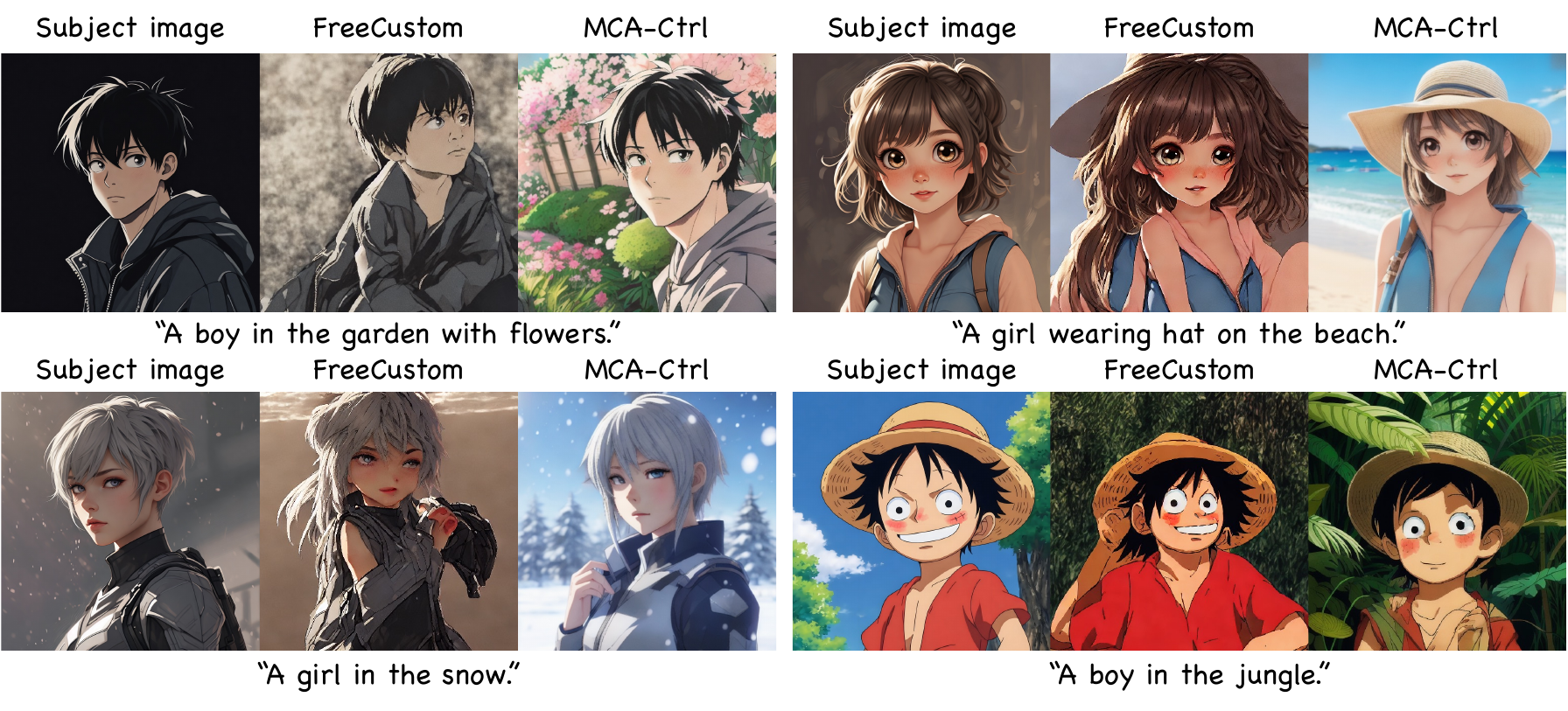}}
  \vspace{-0.2cm}
  \caption{
  Enlarged version of Figure 8 in the main text.
  }
  \label{fig2}
\end{figure*}

\begin{figure*}
\centerline{\includegraphics[width=0.95\linewidth]{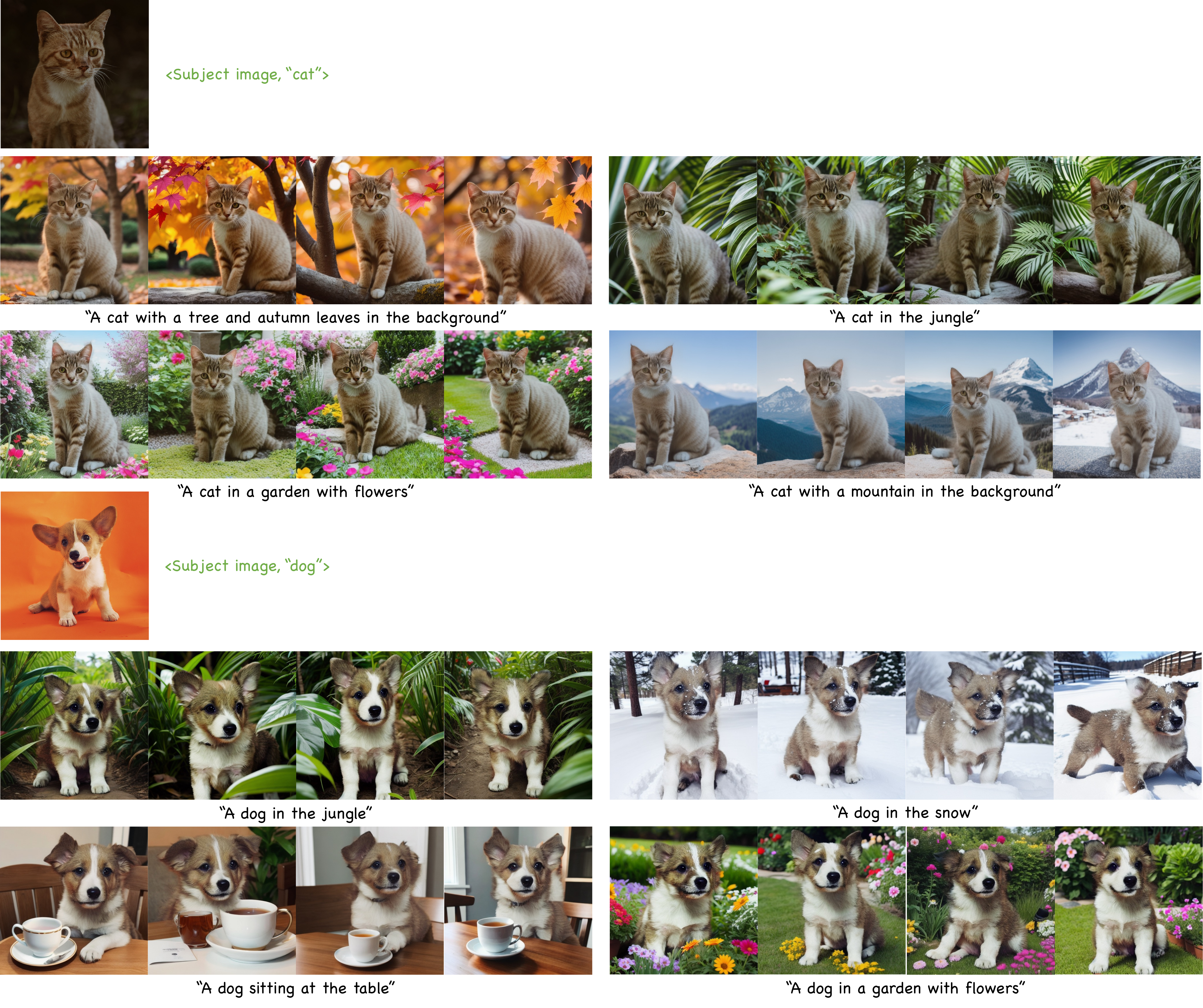}}
\vspace{-0.3cm}
  \caption{
  More customized generation results of MCA-Ctrl (1).
  }
  \label{fig5}
\vspace{-0.5cm}
\end{figure*}

\begin{figure*}
\centerline{\includegraphics[width=1\linewidth]{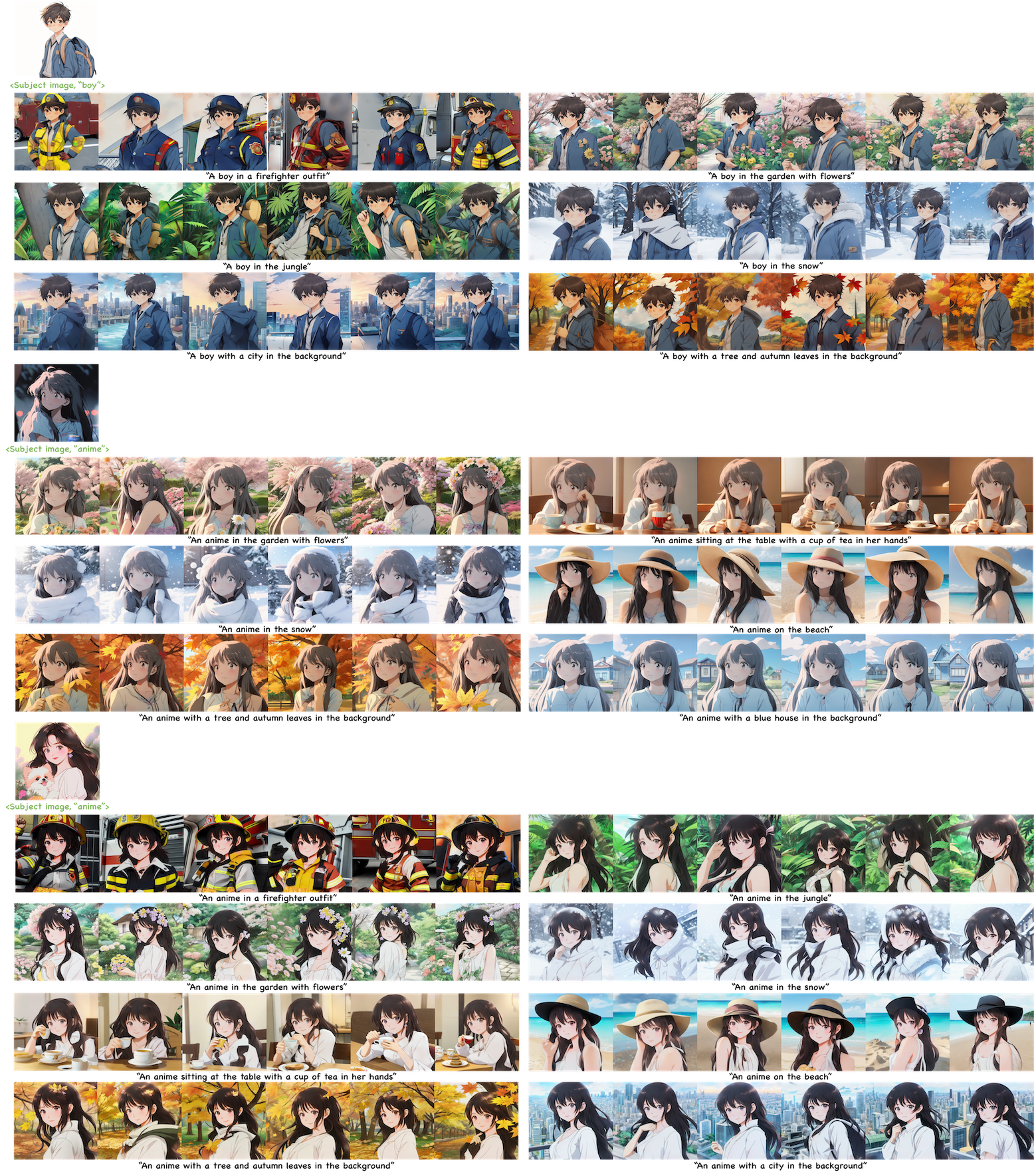}}
  \caption{
  More customized generation results of MCA-Ctrl (2).
  }
  \label{fig4}
\end{figure*}

\begin{figure*}
\centerline{\includegraphics[width=1\linewidth]{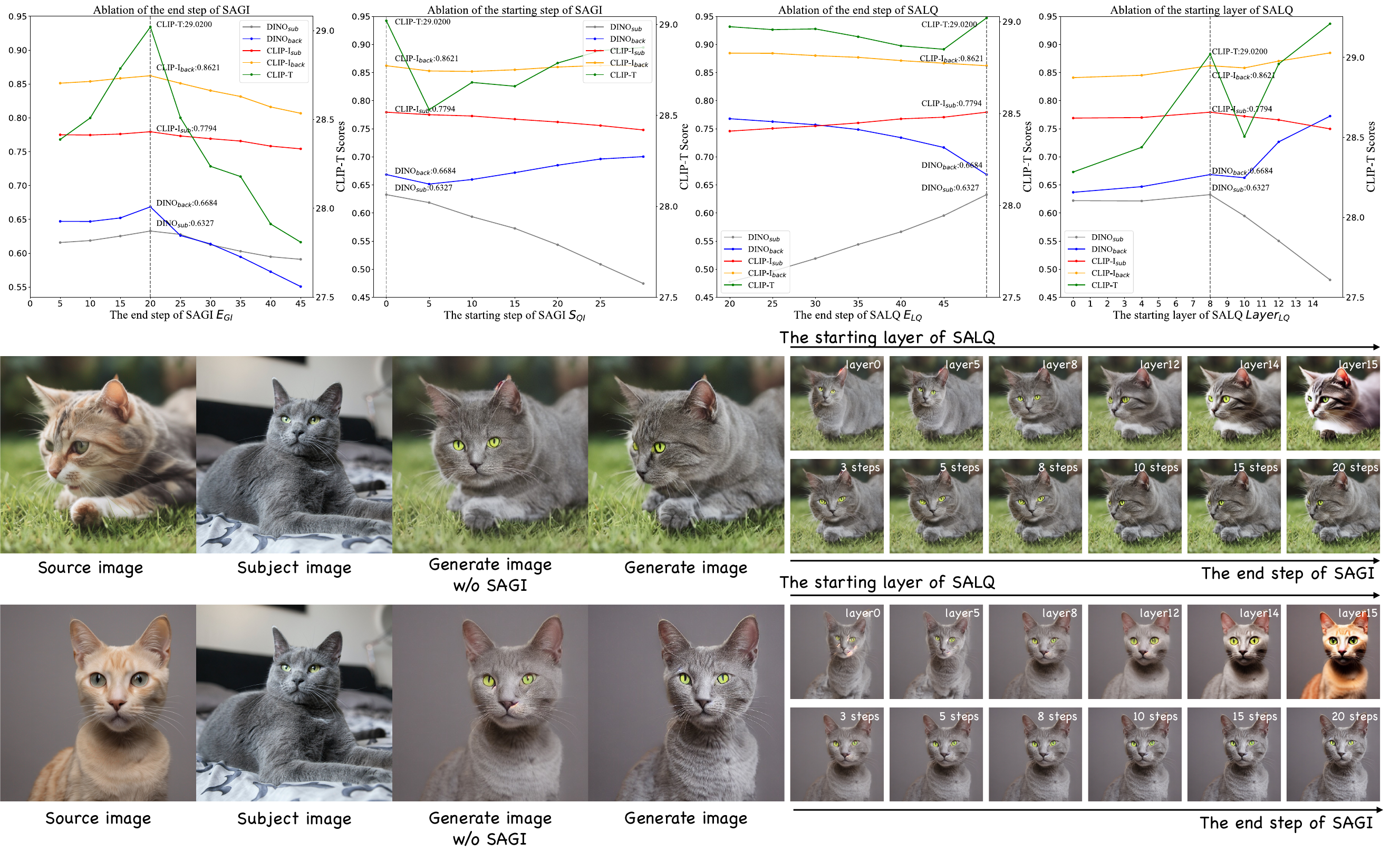}}
\vspace{-0.3cm}
  \caption{
  Enlarged version of Figure 9 in the main text.
  }
  \label{fig3}
  \vspace{-0.3cm}
\end{figure*}



\end{document}